%% file: main.tex
\documentclass{article}

\input{myheaders.tex}

\usepackage{hyperref}
\usepackage{array}
\usepackage{fullpage}
\usepackage{subfigure} 

\title{Reward Models Identify Consistency, Not Causality}
\author{Yuhui Xu\thanks{Email: \texttt{\{yuhui.xu, hanze.dong, leiwang, cxiong, junnan.li\}@salesforce.com}} \qquad Hanze Dong \qquad Lei Wang \\\\ 
 Caiming Xiong \qquad Junnan Li
\\\\
Salesforce AI Research}

\date{}
\usepackage{cancel}

\usepackage[skins]{tcolorbox} 
\tcbuselibrary{breakable} 
\newtcolorbox[auto counter, number within=section, list type=subsubsection, list inside=toc]{sectionbox}[2][]{
colback=white!98!gray, colframe=black, 
colbacktitle=white!90!gray, coltitle=black, 
fonttitle=\bfseries,
title={#2}, 
list entry={Comment \thetcbcounter\quad}
}
\usepackage[utf8]{inputenc}
\begin{document}

\maketitle

\begin{abstract}
Reward models (RMs) play a crucial role in aligning large language models (LLMs) with human preferences and enhancing reasoning quality. Traditionally, RMs are trained to rank candidate outputs based on their correctness and coherence. However, in this work, we present several surprising findings that challenge common assumptions about RM behavior. Our analysis reveals that state-of-the-art reward models prioritize structural consistency over causal correctness. Specifically, removing the problem statement has minimal impact on reward scores, whereas altering numerical values or disrupting the reasoning flow significantly affects RM outputs. Furthermore, RMs exhibit a strong dependence on complete reasoning trajectories—truncated or incomplete steps lead to significant variations in reward assignments, indicating that RMs primarily rely on learned reasoning patterns rather than explicit problem comprehension.  These findings hold across multiple architectures, datasets, and tasks, leading to three key insights: (1) RMs primarily assess coherence rather than true reasoning quality; (2) The role of explicit problem comprehension in reward assignment is overstated; (3) Current RMs may be more effective at ranking responses than verifying logical validity. Our results suggest a fundamental limitation in existing reward modeling approaches, emphasizing the need for a shift toward causality-aware reward models that go beyond consistency-driven evaluation.

\end{abstract}

\setlength{\parindent}{0pt}
\setlength{\parskip}{8pt}
\newpage 
\tableofcontents
\newpage 

\begin{figure}[!ht]
    \centering
    \includegraphics[width=0.9\linewidth]{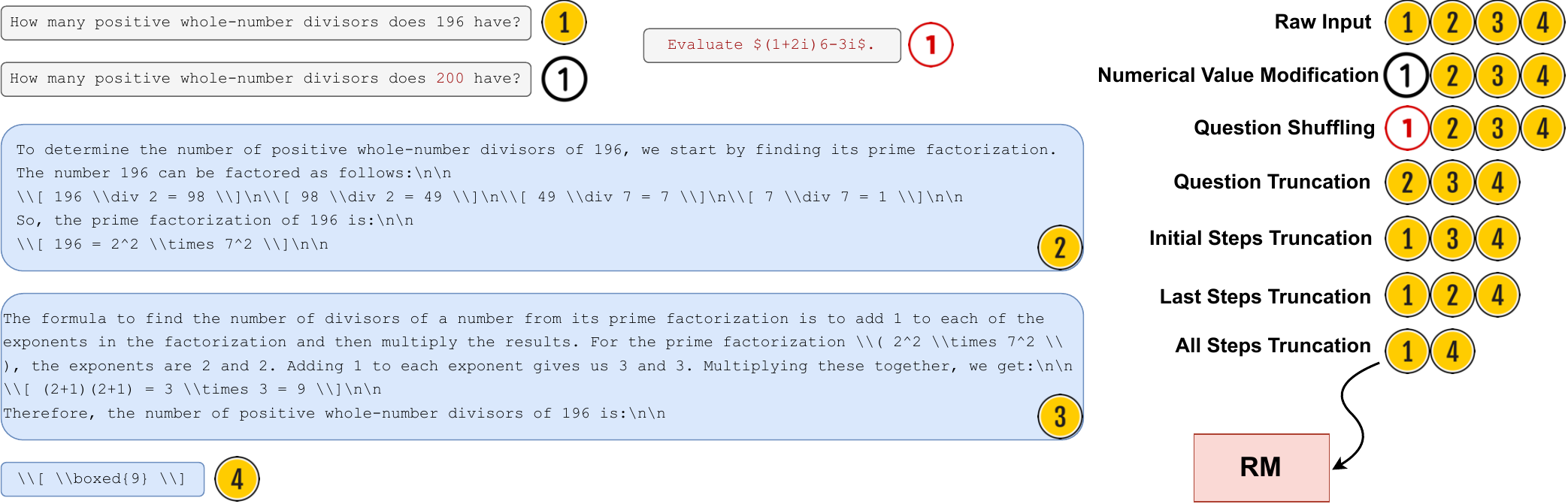}
    \caption{Illustrations of the reward model input modifications. We apply various perturbations, including numerical value modification, question shuffling, question truncation, initial steps truncation, last steps truncation, and all steps truncation, to assess the sensitivity of different input components on reward evaluation.}
    \label{fig:framework}
\end{figure}
\section{Introduction} \label{sec:intro}
Large language models (LLMs)~\citep{hurst2024gpt,dubey2024llama,team2024gemini,anthropic2024claude,jiang2023mistral,liu2024deepseek,yang2024qwen2} have emerged as a dominant paradigm in natural language processing, demonstrating remarkable performance across a diverse range of tasks. The Scaling Law~\citep{kaplan2020scaling} suggests that as model size increases, LLMs develop emergent abilities, enhancing their capacity to comprehend and solve complex tasks. This scalability enables LLMs to generate coherent, contextually accurate responses, supporting a wide array of downstream applications, including summarization~\citep{zhang2019hibert,zhang2024benchmarking}, code generation~\citep{chen2021evaluating}, mathematical reasoning~\citep{hendrycks2021measuring,zhou2023solving}, and conversational AI~\citep{chatgpt,hurst2024gpt}.

A key factor contributing to the success of large language models is their ability to align model outputs with user preferences\citep{christiano2017deep}, which relies on training robust reward models. Beyond preference alignment, reward models also play a crucial role in enhancing reasoning capabilities, serving as mechanisms to evaluate and refine logical correctness in complex tasks\citep{cobbe2021training,lightman2023let}. One promising approach to scaling test-time computation~\citep{snell2024scaling,brown2024large,cobbe2021training,dong2023raft} involves leveraging reward models to search for optimal solutions among multiple candidates. Despite these advancements, the intrinsic mechanisms of reward models remain underexplored—specifically, \textbf{the basis on which they assign rewards to generated trajectories and whether they truly comprehend and reason about the questions they evaluate}. In this paper, we conduct a comprehensive empirical study of state-of-the-art reward models across multiple reasoning datasets and uncover two surprising findings. \textbf{First}, our systematic error analysis (Figure~\ref{fig:framework}) reveals that question truncation has the least impact on reward outputs, whereas modifying numerical values or shuffling the question significantly disrupts reward assignments. This suggests that reward models prioritize internal coherence over true causal understanding—they assess solutions based on structural consistency rather than verifying whether the reasoning directly corresponds to the given question. \textbf{Second}, when provided with incomplete trajectories (i.e., truncated reasoning steps or only given the final answer), the reward outputs change significantly. This indicates that current reward models rely heavily on complete reasoning steps or learned patterns to justify trajectory quality, rather than truly understanding the problem-solving process. Furthermore, our rank correlation analysis and Best-of-$N$ experiments confirm that while reward models remain robust to question omission, they are highly sensitive to the completeness of reasoning steps and the consistency between the question and solution. 

Our results advocate a rethinking of existing reward models. These findings highlight a fundamental limitation of current reward models: they evaluate logical structure rather than verifying causal correctness, raising important questions about their ability to generalize and assess novel problem-solving scenarios effectively.

\section{Related Work} \label{sec:related}

\subsection{LLM Reward Models}
Reward models play a crucial role in human preference alignment~\citep{christiano2017deep,bai2022training,casper2023open} by guiding large language models (LLMs) toward desired behaviors. Broadly, reward modeling methods can be categorized into two approaches.
The first is the preference-based reward model, such as Bradley-Terry (BT) model~\citep{bradley1952rank,zhao2023slic,rafailov2024direct,ethayarajh2024kto,xiong2024iterative} and general preference model \citep{llm-blender-2023,munos2023nash,tang2024generalized,ye2025online,azar2024general}, which defines the reward function by the preference between two responses.
Conventional RLHF usually capture the human preference with BT model \citep{ouyang2022training,chatgpt}, which has been widely proven to improve the quality of model outputs \citep{dubey2024llama,dong2024rlhf,guo2024direct}.
The second approach estimate the probability of correctness as rewards, directly scoring outputs without relying on pairwise comparisons. 
In this paper, we primarily focus on correctness-based cases, which are well-defined and more commonly used for selecting reasoning trajectories during both training~\citep{chen2024alphamath,wang2024math} and inference~\citep{brown2024large} in reasoning tasks. Depending on how reward signals are assigned, these models can be classified into Outcome Reward Models (ORMs) and Process Reward Models (PRMs). ORMs~\citep{yu2023outcome} evaluate solutions based solely on the final output, while PRMs~\citep{lightman2023let} provide step-level annotations, offering dense and granular reward signals at each reasoning step to encourage structured problem-solving. PRMs have been proven to be effective in mathematical problems \citep{shao2024deepseekmath,snell2024scaling,luo2024improve,liao2025reward}. 

\subsection{Robustness of Reward Models}
Despite the success of reward models in aligning with human preferences, they still have issues. A common issue is reward hacking~\citep{ibarz2018reward, denison2024sycophancy}, where the policy achieves high reward scores from the reward model without exhibiting the desired behavior. This phenomenon leads to performance degradation~\citep{bai2022training} and increases the discrepancy between the policy model's behavior and the intended objective~\citep{stienon2020learning}. 
Reward hacking manifests in various patterns~\citep{park2024offsetbias}, with length hacking being one of the most prevalent and well-documented cases in large language model research.~\citet{singhal2024long} investigate length-related issues in reward models, demonstrating a strong correlation between reward scores and text length. This finding aligns with the observation by~\citet{dubois2023alpacafarm} that output length distributions tend to increase after applying PPO. And~\citet{liu2024iterative} explore length hacking with the popular DPO algorithm. In addition, ODIN~\citep{denison2024sycophancy} explores to mitigate the length hacking issue by disentangling the length from the original reward.
In this work, rather than exploring new general patterns of reward hacking or developing mitigation techniques, we focus on an empirical study that explores whether state-of-the-art reward models genuinely understand questions, reasoning steps, and their causal relationships in reasoning tasks.

\section{LLM Reward}

We formalize the interaction between a user and an LLM as a mapping from a given context or prompt, denoted as \( x \), to a generated response \( y \). The response \( y \) consists of a sequence of steps, represented as \( y = [y_1, \dots, y_n] \). The ideal reward function \( r_*(x, y) \) quantifies the quality of the generated response in terms of its final performance. To ensure a well-defined reward function, we adopt an outcome-based formulation and focus on objective reasoning problems, where the reward is defined as the probability that \( y \) produces a correct or desirable outcome:
\[
r_*(x, y) = \textbf{P}(y\text{ is correct} \mid x).
\]
This probabilistic formulation provides a structured measure of response effectiveness. By framing the reward in this manner, we ensure that learning objectives align with producing accurate and reliable responses.

In real world, although we can access the reward for training data. People usually use another LLM to estimate the optimal reward $r_\theta$. Ideally, $r_\theta  \approx r_*$.
To train a reward model, people use a dataset of labeled examples \(\{(x_i, y_i, z_i)\}_{i=1}^n\), where \( x_i \) represents the input prompt, \( y_i \) is the generated response, and \( z_i \in \{1, 0\} \) is a binary label indicating whether the response is correct (\(1\)) or incorrect (\(0\)). The reward model \( r_\theta(x, y) \) is parameterized by \(\theta\) and is trained to approximate the ideal reward function \( r_*(x, y) \) by minimizing a loss function that encourages consistency with the labeled data. A common approach is to optimize a binary cross-entropy loss:
\[
\mathcal{L}(\theta) = -\sum_{i=1}^n \big[ z_i \log r_\theta(x_i, y_i) + (1 - z_i) \log (1 - r_\theta(x_i, y_i)) \big],
\]

where \( r_\theta(x_i, y_i) \) is interpreted as the probability that \( y_i \) is correct given \( x_i \). This formulation ensures that the reward model learns to distinguish between high-quality and low-quality responses. Once trained, the reward model can be used to guide response generation in reinforcement learning or ranking-based optimization frameworks.

In addition, people are refining the reward model with the process-based supervision.
$$
r_*(x,y_{1:k}) = \textbf{P}(y_{1:k}\text{ is correct} \mid x).
$$
By incorporating process-based supervision, reward models can capture fine-grained signals that improve alignment with human reasoning, ultimately leading to more interpretable and controllable LLM outputs. This paradigm enables reward models to provide more structured feedback, particularly for tasks requiring multi-step reasoning or sequential decision-making.

In this paper, we would like to investigate the reward behavior of incomplete inputs to identify what are really matters for reward models. For example, empty input 
 $r_\theta(\mathtt{None},y)$, truncated output $r_\theta(x,y_{1:n/2})$ or $r_\theta(x,y_{n/2:n})$, shuffled input $r_\theta(x',y)$.

\begin{figure}[!ht]
    \centering
    \subfigure[MATH500]{
        \includegraphics[width=0.45\textwidth]{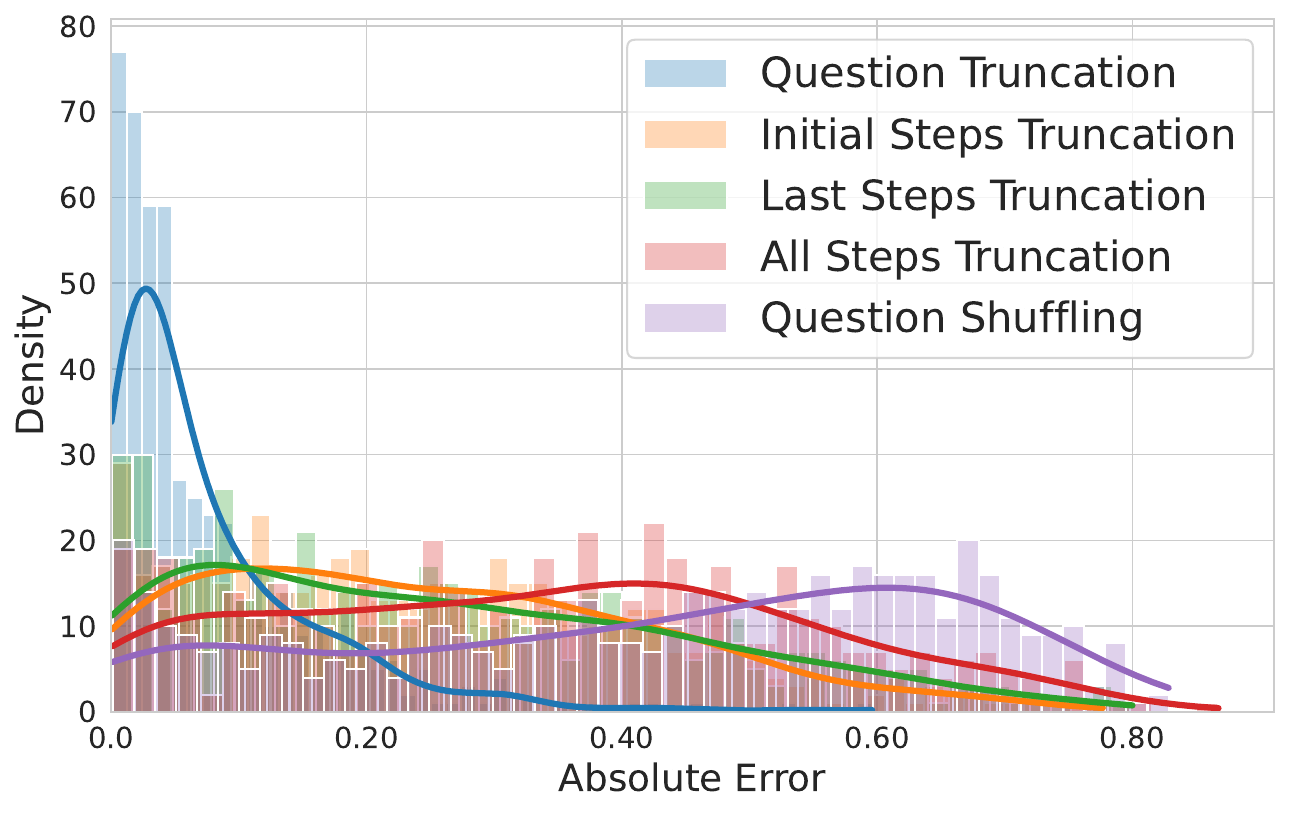}
    }
    \subfigure[GSM8K]{
        \includegraphics[width=0.45\textwidth]{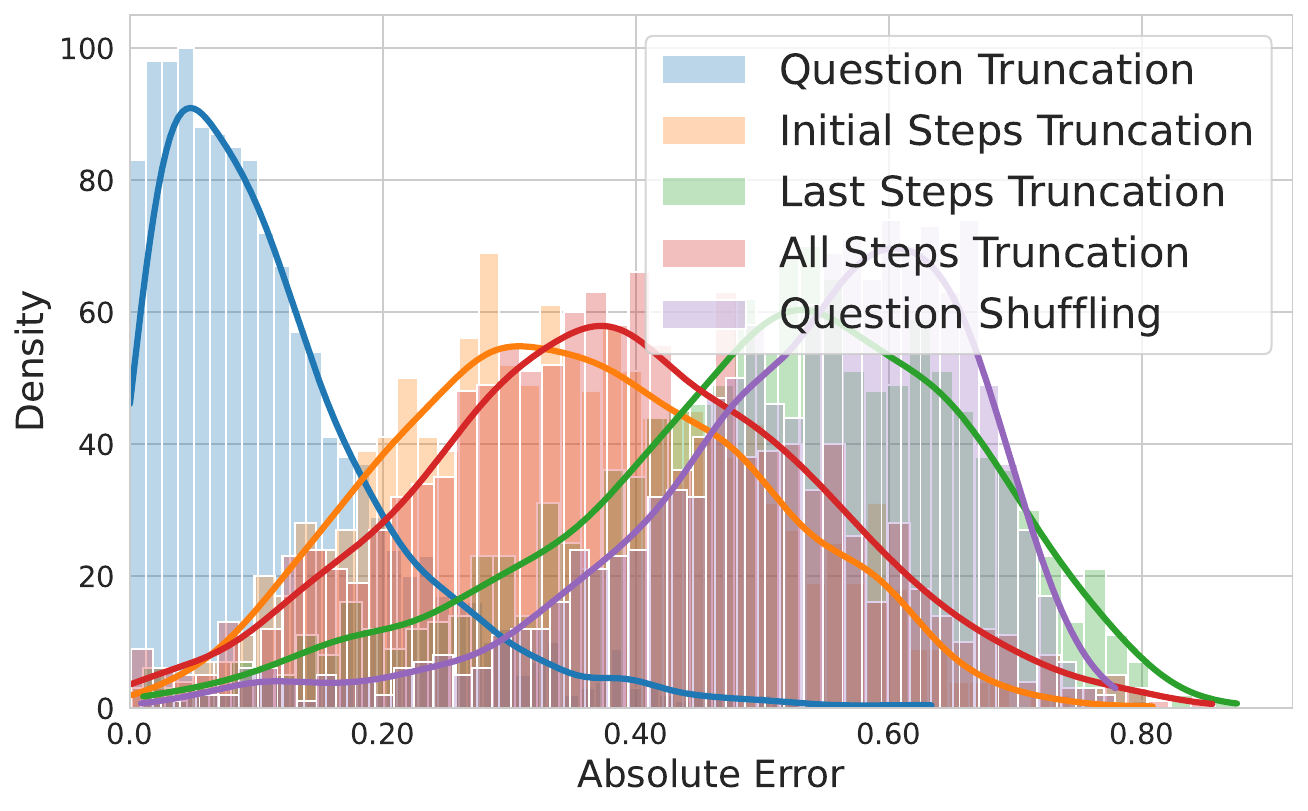}
    }
    
    \subfigure[Olympiad Bench]{
        \includegraphics[width=0.45\textwidth]{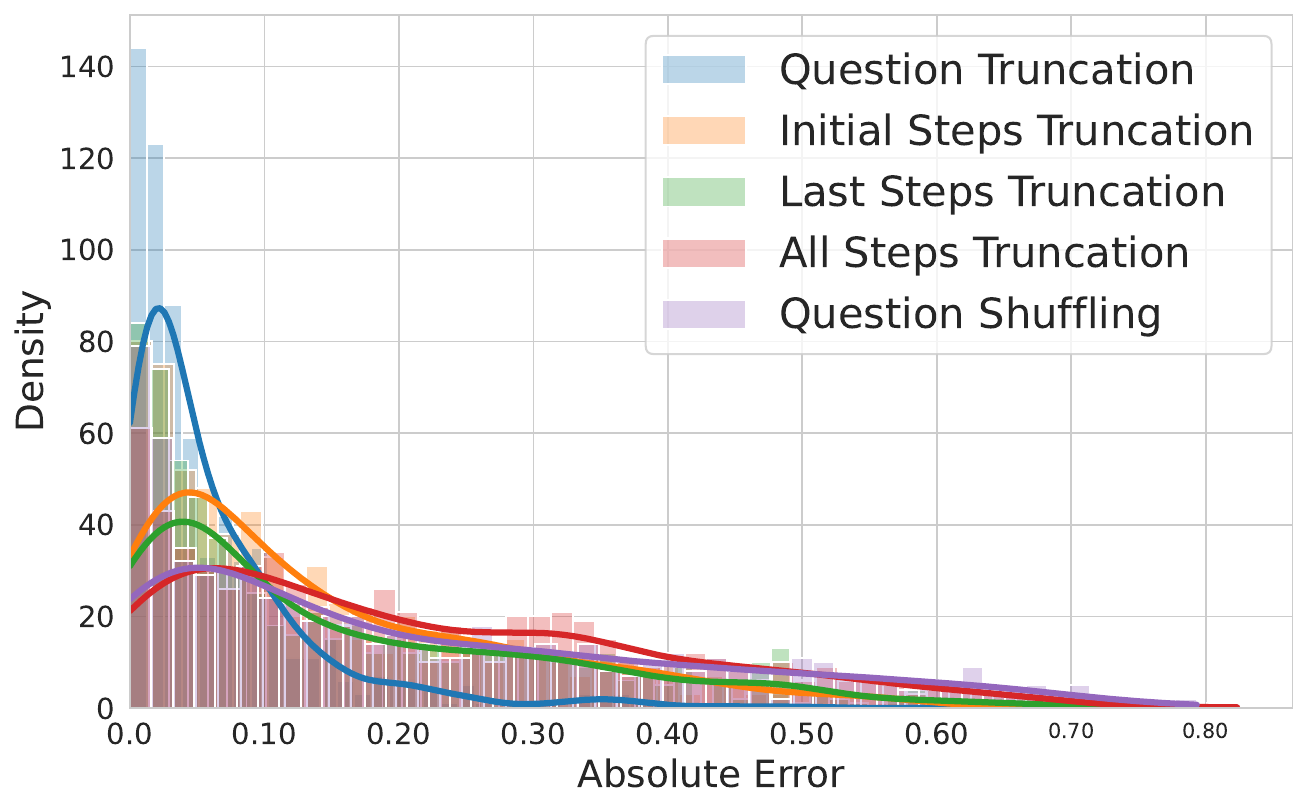}
    }
    \subfigure[GaoKao-2023-En]{
        \includegraphics[width=0.45\textwidth]{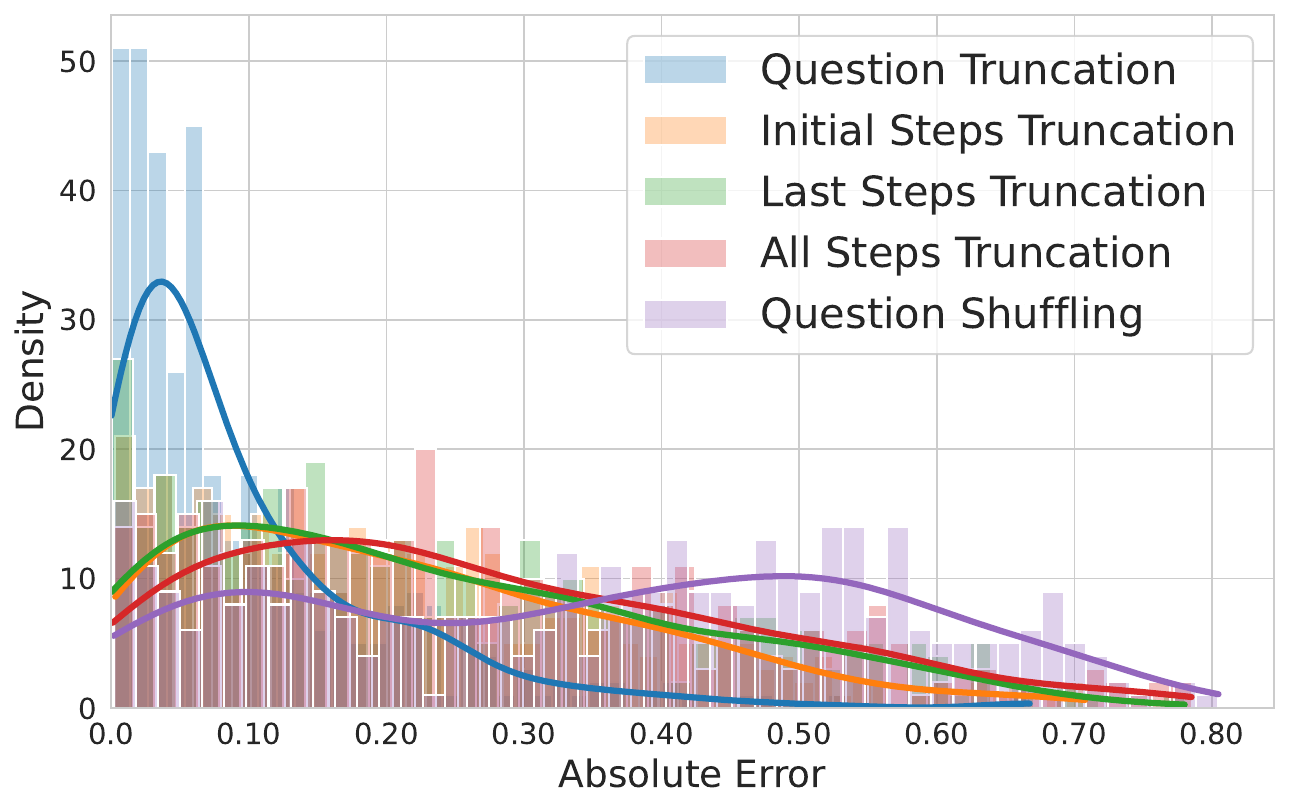}
    }
    
    \caption{Density maps of absolute reward errors across different datasets when truncating various input components, using Qwen-2.5-Math-1.5B-Instruct as the base model and Skywork-O1-Open-PRM-Qwen-2.5-1.5B as the reward model..}
    \label{fig:truncation}
\end{figure}

\section{Experimental Setup} \label{sec:setup}
We outline the experimental setup used in our analysis in Sections~\ref{sec:error} and \ref{sec:bon}.

\paragraph{Models.} We utilize two reward model families: Skywork-o1-OpenPRM~\citep{skyworkopeno12024} and RLHFlow~\citep{xiong2024implementation}. Specifically, our experiments include Skywork-o1-Open-PRM-Qwen-2.5-1.5B, Skywork-o1-Open-PRM-Qwen-2.5-7B, Llama3.1-8B-ORM-Deepseek-Data, and Llama3.1-8B-PRM-Deepseek-Data. For base models, we employ both general-purpose and math-focused LLMs, specifically Llama-3~\citep{dubey2024llama} and Qwen-2.5-Math~\citep{yang2024qwen2math}. 

\paragraph{Datasets.} We conduct experiments on a diverse set
of reasoning tasks, including GSM8K~\citep{cobbe2021training}, MATH500~\citep{hendrycks2021measuring}, OlympiadBench~\citep{he2024olympiadbench}, GaoKao-2023-En~\citep{liao2024mario}, and Minerva Math~\cite{minerva}.

\paragraph{Default Setting.} All experiments were conducted on NVIDIA H100 GPUs, using vLLM\citep{kwon2023efficient} as the backend. We set the generation parameters to \texttt{temperature = 0.8} and \texttt{top\_p = 1.0} for Best-of-$N$ sampling and trajectory collection. A reasoning step is defined as a generation ending with \texttt{\textbackslash n\textbackslash n}. For the process reward model, each trajectory is scored based on the reward of its final step. In the error analysis experiments (Section\ref{sec:error}), we sample 32 trajectories per question.

\section{Questions Matters Little} \label{sec:error}
\subsection{Which Input Matters Most?}
To assess the relative importance of different components in reward model inputs, we systematically truncate various parts of the input and analyze their impact on model predictions. Specifically, we evaluate how the absence of key information—such as the question or portions of the solution—affects the reward assignment. The following truncation strategies are considered:

\textbf{Question Truncation}: The question is entirely removed, leaving only the solution trajectory as input to the reward model. This tests whether the model primarily relies on the reasoning process rather than the problem statement itself.

\textbf{Initial Steps Truncation}: The first half of the solution steps is removed, preserving only the latter portion. This examines whether early-stage reasoning contributes significantly to the model’s assessment or if later steps alone are sufficient.

\textbf{Last Steps Truncation}: The final half of the solution steps is removed, retaining only the initial portion. This helps determine whether the model emphasizes intermediate reasoning steps or prioritizes the final stages of problem-solving.

\textbf{All Steps Truncation}: All solution steps are removed, leaving only the final answer box. This scenario isolates the influence of the final answer on the reward model’s judgment, shedding light on whether the model evaluates reasoning quality or focuses primarily on correctness.

We quantify the impact of input modifications by computing the absolute error between the original reward output, $r$, and the reward of the truncated input, $r^*$, given by $|r - r^*|$. The results on MATH500, GSM8K, OlympiadBench, and GaoKao-2023-En are visualized in Figure~\ref{fig:truncation}. We present histograms of the error distribution across all questions, with each question sampled over 32 trajectories.

We observe that the error distributions vary across datasets and truncation strategies, highlighting the differing sensitivities of reward models to different input components. Question truncation generally results in a lower absolute error, suggesting that the model can still assign reasonable reward scores based on the solution trajectory alone. In contrast, initial steps truncation and last steps truncation exhibit distinct effects, with the latter often leading to larger errors, implying that final reasoning steps play a crucial role in the model’s decision-making process. The all steps truncation condition, which retains only the final answer box, consistently produces the highest errors, particularly in MATH500 and OlympiadBench, indicating that intermediate reasoning steps are essential for accurate reward assignment.

These findings suggest that the question is less important than the reasoning steps in determining reward scores. The relatively low absolute error from question truncation indicates that the model can still evaluate solution quality even without explicit access to the problem statement. Conversely, the higher errors observed in last steps truncation and all steps truncation highlight the critical role of intermediate and final reasoning steps in the model’s decision-making process. This suggests that reward models prioritize logical coherence and solution completeness over simply understanding the original question, emphasizing the necessity of reasoning depth in reward evaluation.

\subsection{Consistency Matters}
\begin{figure}[t]
    \centering
    \subfigure[MATH500]{
        \includegraphics[width=0.45\textwidth]{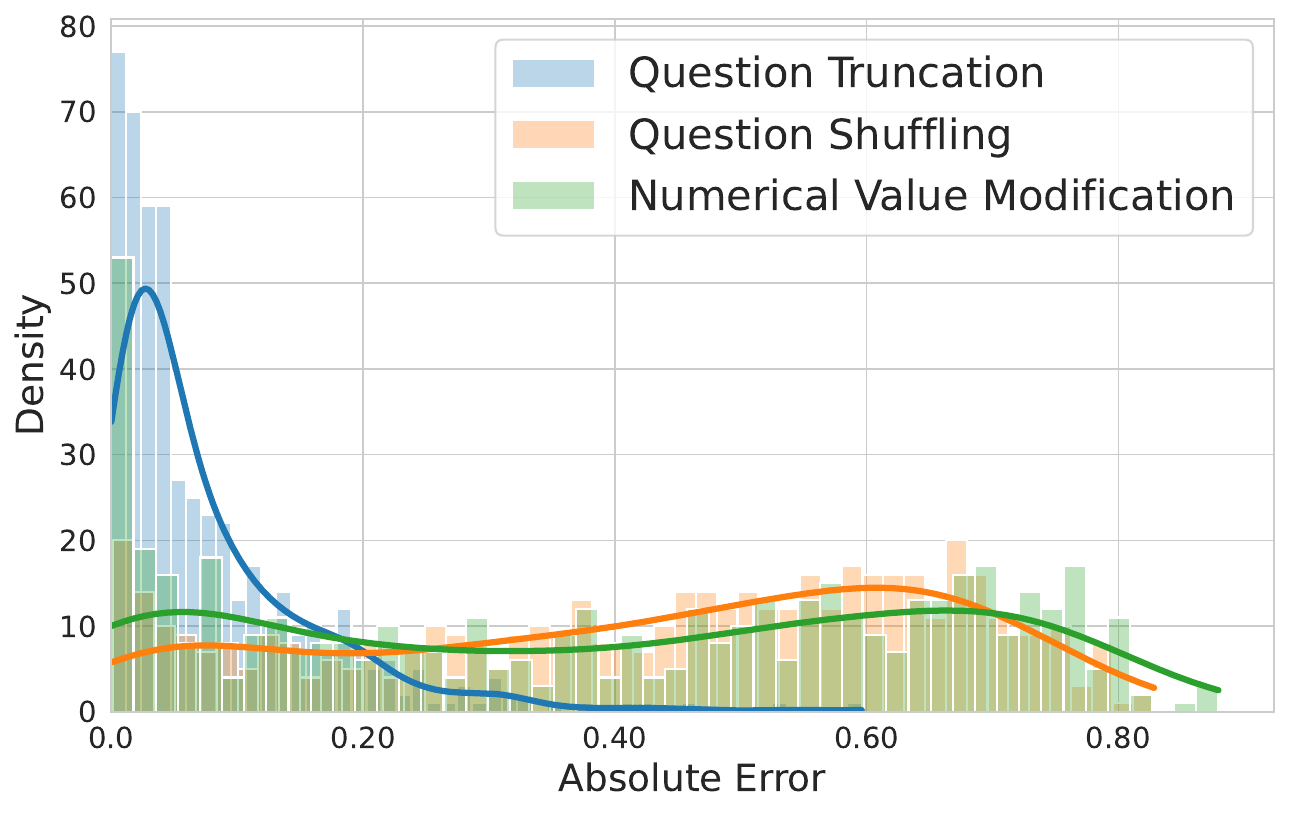}
    }
    \subfigure[GSM8k]{
        \includegraphics[width=0.45\textwidth]{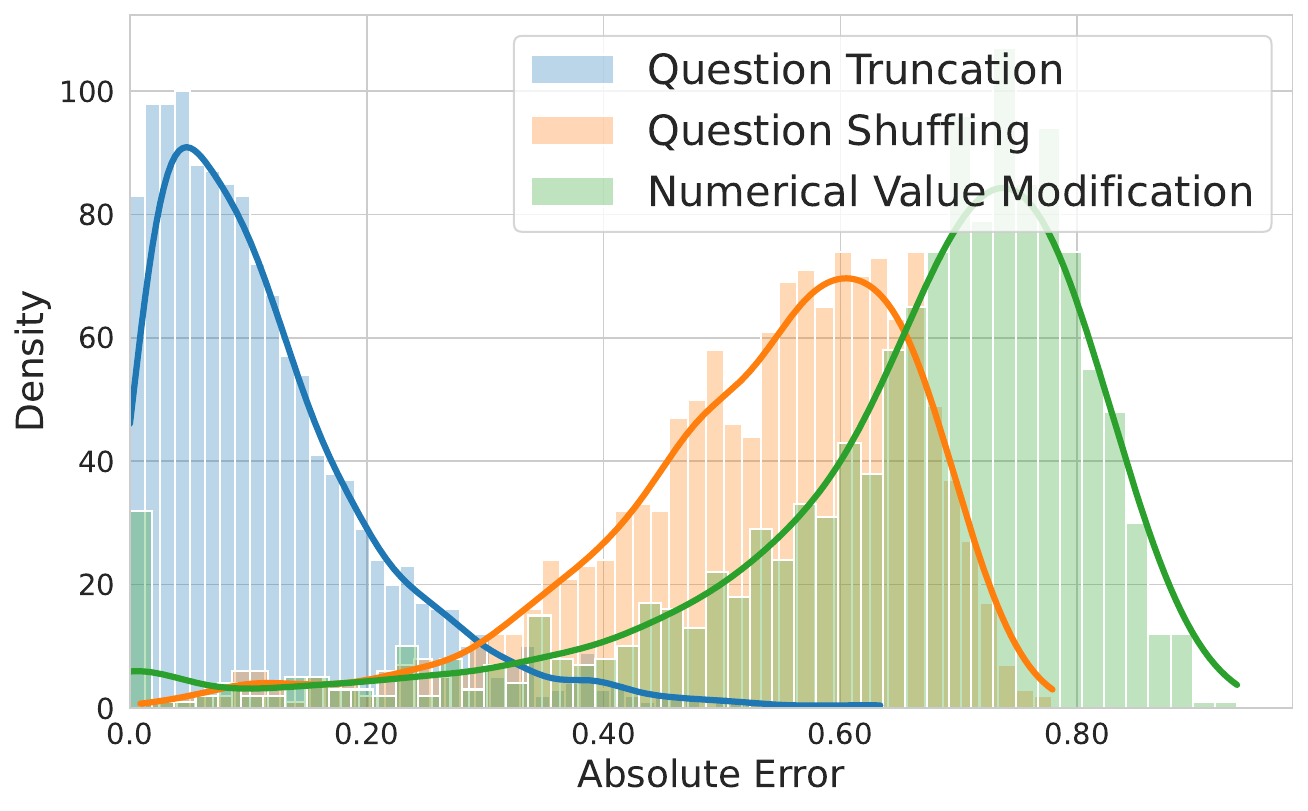}
    }
    
    \subfigure[Olympiad Bench]{
        \includegraphics[width=0.45\textwidth]{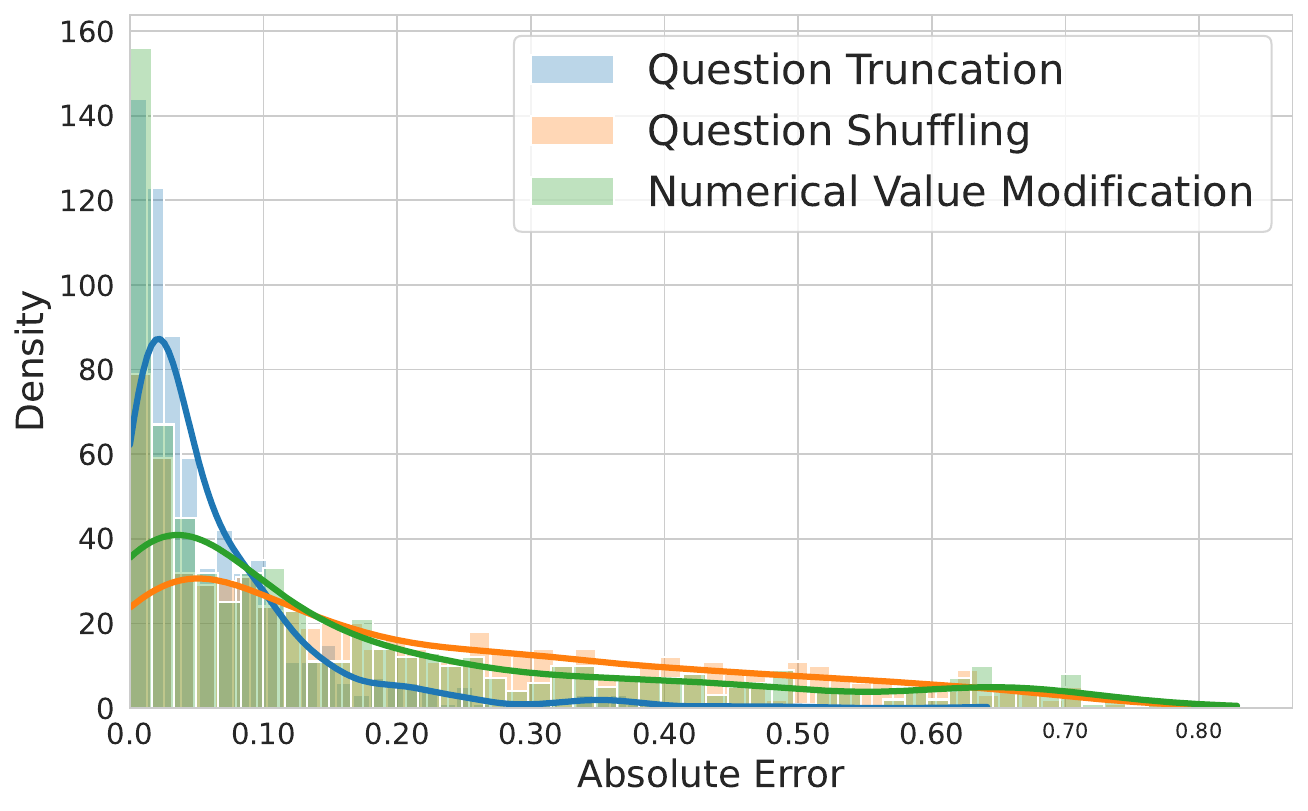}
    }
    \subfigure[Gaokao2023 En]{
        \includegraphics[width=0.45\textwidth]{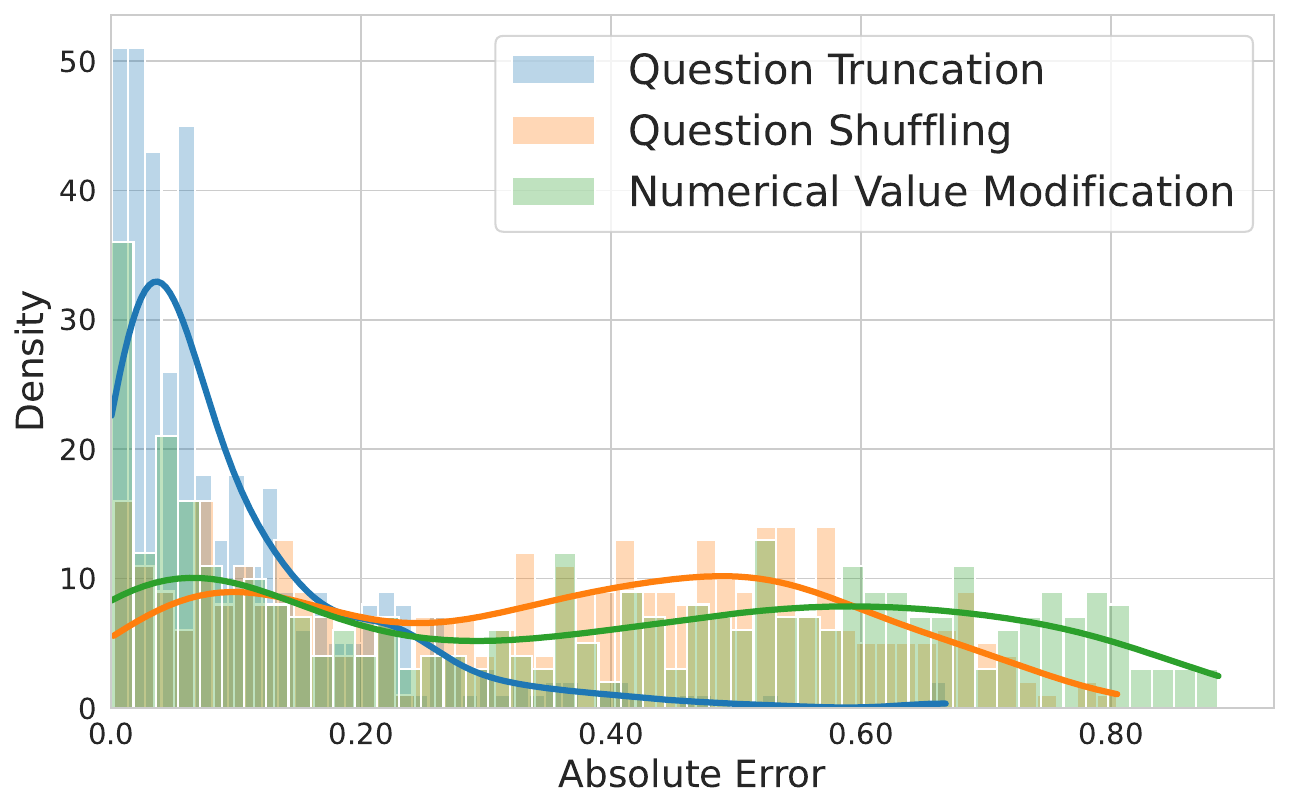}
    }
    
    \caption{Density maps of absolute reward errors across different datasets. We compare the effects of question truncation, question shuffling, and numerical value modification using Qwen-2.5-Math-1.5B-Instruct as the base model and Skywork-O1-Open-PRM-Qwen-2.5-1.5B as the reward model.}
    \label{fig:number}
\end{figure}
To gain deeper insights into the role of the question in reward modeling, we conduct a series of controlled experiments designed to assess the extent to which the reward model relies on the problem statement for evaluating solution quality. Specifically, we investigate how disrupting the question-solution relationship and altering key numerical values affect the model’s reward assignment.

\textbf{Question Shuffling}: We shuffle the questions and their corresponding solution trajectories, disrupting the original question-solution pairings. This tests the reward model’s reliance on the semantic coherence between the problem statement and the reasoning steps.

\textbf{Numerical Value Modification}: We replace numerical values in the question with random values, altering the problem while preserving its overall structure. This evaluates the model’s sensitivity to specific numerical details in the question.

Figure~\ref{fig:number} presents the absolute error distributions for different question modification strategies across four reasoning benchmarks. Across all datasets, question truncation consistently results in lower errors, suggesting that removing the question has a limited impact on the reward model. Question shuffling introduces moderate errors, particularly in GSM8K, indicating that the semantic consistency between the problem and solution plays a role in reward assignment. Numerical value modification, especially in GSM8K, leads to the largest errors, demonstrating that changes in numerical details significantly affect the reward model’s predictions.

These results highlight the importance of consistency in the reward model’s input. The fact that question truncation results in lower errors implies that the reward model primarily evaluates the reasoning process rather than the question itself. However, the increased errors in question shuffling suggest that disrupting semantic alignment between the problem and solution trajectory weakens the model's ability to assign consistent rewards. Moreover, the high error from numerical value modification indicates that numerical consistency is crucial, as the model relies on specific values to gauge correctness.

\begin{figure}[t]
    \centering
    \subfigure[MATH500]{
        \includegraphics[width=0.45\textwidth]{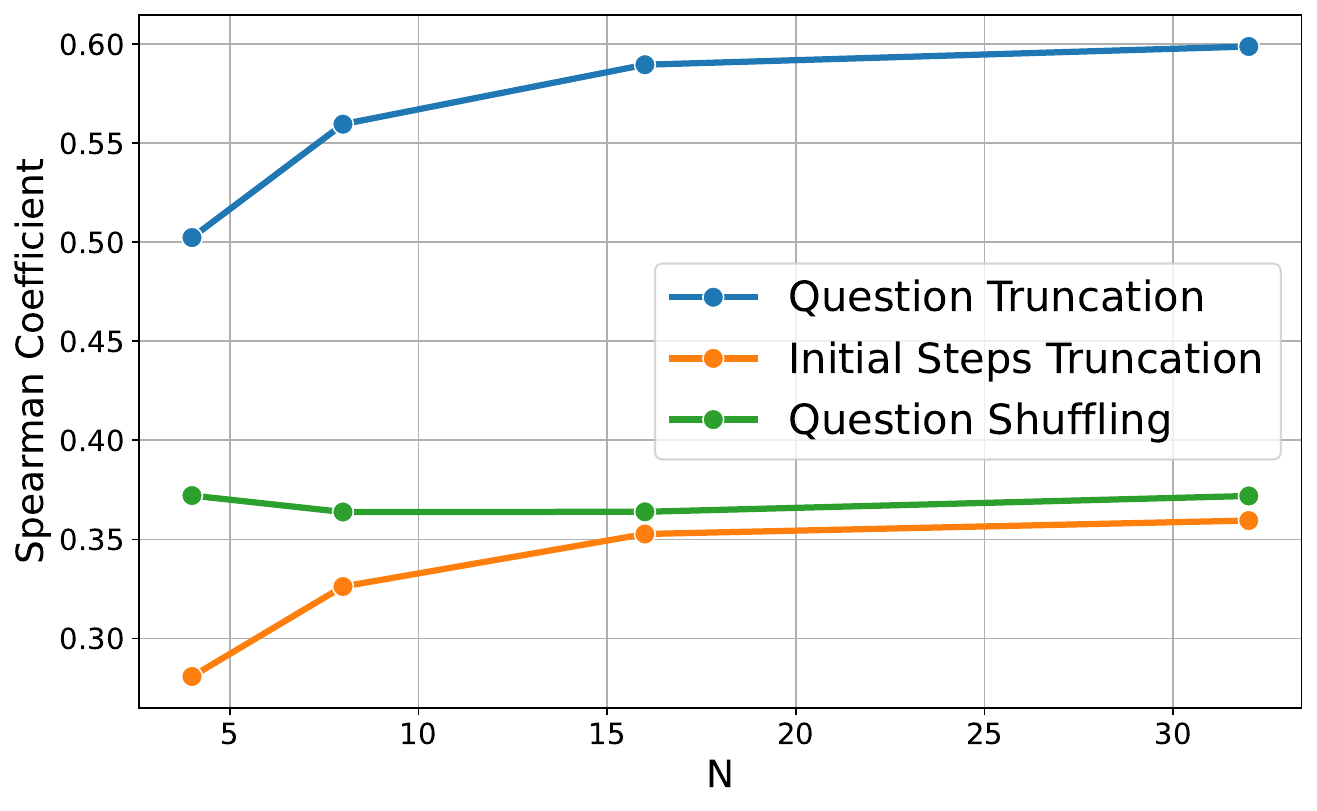}
    }
    \subfigure[Olympiad Bench]{
        \includegraphics[width=0.45\textwidth]{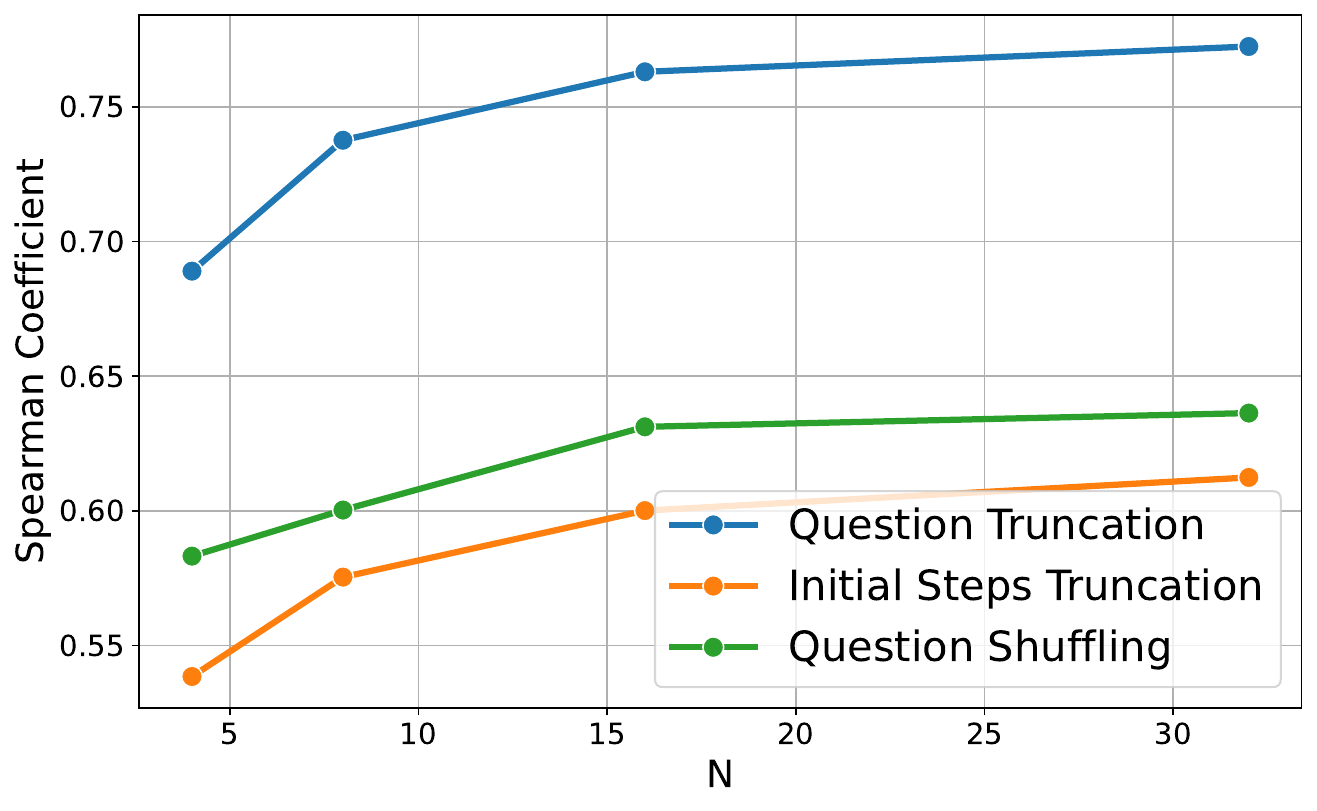}
    }

    \caption{Spearman correlation of reward rankings under input truncation and Shuffling on MATH500 and Olympiad Bench. $N$ is the number of sampled trajectories for each input question.}
    \label{fig:coeff}
\end{figure}

\begin{table*}[!ht]
\caption{Best-of-$N$ results under different reward input modifications on Skywork-o1-Open-PRM.}
\label{tab:qwen}
\vskip 0.1in
\scriptsize
\begin{center}
\begin{tabular}{crlcccccl}
\toprule
\multirow{2}{*}{\textbf{Base Model}} & \multirow{2}{*}{\textbf{RM}} & \multirow{2}{*}{\textbf{Setting}} & \multirow{2}{*}{\textbf{MATH500}} & \multirow{2}{*}{\textbf{GSM8K}} &  \textbf{Olympiad} &  \textbf{Minerva} & \multirow{2}{*}{\textbf{Avg.}} \\
 & & & &  &  \textbf{Bench} &  \textbf{Math} \\
\midrule
7B & - & - & 83.2 & 95.7 & 41.2 & 37.5 &  64.4\\
\midrule
\multicolumn{7}{c}{\textit{Math Model, Base: Qwen2.5-Math-Instruct, RM: Skywork-
o1-Open-PRM 1.5B}} \\
\midrule
 7B & 1.5B & $N=4$ & 85.4 & 96.9 & 43.3 & 39.3 & 66.2\\
 7B & 1.5B & $N=8$ & 84.6 & 96.7 & 45.3 & 38.2 & 66.2\\
 7B & 1.5B & $N=16$ & 85.4 & 96.9 & 45.0 & 38.6 & 66.5\\
\cmidrule(lr){1-8}
\multicolumn{8}{c}{\textit{Question Truncation}} \\
 7B & 1.5B & $N=4$ & 85.8 & 96.4 & 43.0 & 36.0 & 65.3\\
 7B & 1.5B & $N=8$ & 86.4 & 97.0 & 42.7 & 37.9& 66.0\\
 7B & 1.5B & $N=16$ & 86.0 & 96.1 & 45.3 & 37.1 & 66.1\\
\multicolumn{8}{c}{\textit{Initial Steps Truncation}} \\
 7B & 1.5B & $N=4$ & 85.0 & 96.4 & 39.7 & 36.4 & 64.4\\
 7B & 1.5B & $N=8$ & 86.4 & 96.3 & 43.0 & 36.4& 65.5\\
 7B & 1.5B & $N=16$ & 85.4 & 96.0 & 43.1 & 37.5 & 65.5\\
\multicolumn{8}{c}{\textit{Last Steps Truncation}} \\
 7B & 1.5B & $N=4$ & 84.0 & 95.5 & 42.4 & 38.2 & 65.0\\
 7B & 1.5B & $N=8$ & 87.0 & 96.1 & 38.8 & 34.9 & 64.2\\
 7B & 1.5B & $N=16$ & 83.6	& 95.6 & 41.8 & 37.5 & 64.6\\
\multicolumn{8}{c}{\textit{Question Shuffling}} \\
 7B & 1.5B & $N=4$ & 84.8 & 95.9 & 41.9 & 37.5 & 65.0\\
 7B & 1.5B & $N=8$ & 85.0 & 96.1 & 42.7 & 37.1 & 65.2\\
 7B & 1.5B & $N=16$ & 85.0 & 96.7 & 44 & 36.0 & 65.4\\
\midrule
\multicolumn{7}{c}{\textit{Math Model, Base: Qwen2.5-Math-Instruct, RM: Skywork-
o1-Open-PRM 7B}} \\
\midrule
 7B & 7B & $N=4$ & 86.0 & 97.0 & 41.8 & 38.2 & 65.8\\
 7B & 7B & $N=8$ & 87.0 & 97.1 & 45.3 & 37.5& 66.7\\
 7B & 7B & $N=16$ & 86.0 & 97.1 & 47.0 & 39.3 & 67.4\\
\cmidrule(lr){1-8}
\multicolumn{8}{c}{\textit{Question Truncation}} \\
 7B & 7B & $N=4$ & 86.2 & 96.4 & 44.0 & 35.3 & 65.5\\
 7B & 7B & $N=8$ & 86.0 & 96.2 & 43.1 & 37.9 & 65.8\\
 7B & 7B & $N=16$ & 84.0 & 96.3 & 46.5 & 41.9 & 67.2\\
\multicolumn{8}{c}{\textit{Initial Steps Truncation}} \\
 7B & 7B & $N=4$ & 82.8 & 96.3 & 41.5 & 38.2 & 64.7\\
 7B & 7B & $N=8$ & 85.4 & 96.7 & 42.7 & 40.1 & 66.2\\
 7B & 7B & $N=16$ & 84.6 & 96.3 & 43.9 & 40.8 & 66.4\\
\multicolumn{8}{c}{\textit{Last Steps Truncation}} \\
 7B & 7B & $N=4$ & 84.4 & 96.4 & 39.9 & 39.7 & 65.1\\
 7B & 7B & $N=8$ & 83.4 & 96.1	& 40.7 & 37.1 & 64.3\\
 7B & 7B & $N=16$ & 85.6 & 96.4 & 43 & 37.1 & 64.5\\
\multicolumn{8}{c}{\textit{Question Shuffling}} \\
 7B & 7B & $N=4$ & 83.4 & 96.6 & 39 & 34.9 & 63.5\\
 7B & 7B & $N=8$ & 81.8 & 95.8 & 43 & 36.8 & 64.4\\
 7B & 7B & $N=16$ & 83.2 & 95.3 & 41.6	& 37.9 & 64.7\\
\bottomrule
\end{tabular}
\end{center}
\vskip -0.2in
\end{table*}
\section{Impact on Ranking of Rewards}\label{sec:bon}
A key application of reward models is to rank candidate outputs and select the best among them. Beyond absolute error analysis, it is crucial to evaluate how input modifications influence the relative ranking of rewards assigned to different outputs. To this end, we investigate the impact of question and reasoning modifications on ranking consistency through two complementary analyses:
\begin{itemize}
    \item Rank Correlation of Rewards: We assess how well the reward model preserves the relative ranking of outputs after input modifications by computing ranking correlation metrics.
    \item Best-of-$N$ Selection: We evaluate the effect of input modifications on Best-of-$N$ selection performance, where the reward model is used to choose the best candidate from a set of generated outputs.
\end{itemize}
\subsection{Rank Correlation of Rewards}
Using Qwen2.5-Math-7B-Instruct as the base model and Skywork-o1-Open-PRM-Qwen-2.5-1.5B as the reward model, we generate 4, 8, 16, and 32 candidate outputs per question on MATH500 and OlympiadBench. To assess the stability of reward rankings under input modifications, we compute Spearman's rank correlation coefficient ($\rho$) as our ranking consistency metric. Spearman’s correlation quantifies the monotonic relationship between two variables, providing insight into how well the reward model preserves the relative ranking of candidate outputs despite perturbations in input structure. A high Spearman correlation indicates that the ranking of outputs remains stable regardless of modifications. Conversely, a low correlation signifies that input modifications significantly alter the ranking behavior, revealing potential inconsistencies in the model’s reward assignment process.

Figure~\ref{fig:coeff} presents the Spearman rank correlation coefficients for different truncation strategies across varying values of 
$N$. Removing the question consistently results in the highest rank correlation across all values of 
$N$, suggesting that the reward model remains relatively stable in ranking outputs even when the problem statement is absent. Additionally, the correlation increases as $N$ grows, indicating that the ranking consistency improves when selecting from a larger set of candidate outputs. In contrast, removing the first half of the solution steps leads to a significantly lower correlation, implying that complete reasoning steps is important in determining rankings. Furthermore, disrupting the semantic coherence between the question and solution has a noticeable impact on ranking correlation, demonstrating that maintaining a logically consistent input structure is important for stable ranking performance.

These findings further confirms that reward models prioritize reasoning consistency over direct question comprehension. While the models demonstrate robustness to question removal, their sensitivity to reasoning disruptions suggests that they rely more on structural coherence than an actual understanding of problem-solving logic.

\begin{table*}[!t]
\caption{Best-of-$N$ results under different reward input modifications on Llama3.1-8B-ORM/PRM.}
\label{tab:llama}
\vskip 0.1in
\scriptsize
\begin{center}
\begin{tabular}{crlcccccl}
\toprule
\multirow{2}{*}{\textbf{Draft}} & \multirow{2}{*}{\textbf{RM}} & \multirow{2}{*}{\textbf{Setting}} & \multirow{2}{*}{\textbf{MATH500}} & \multirow{2}{*}{\textbf{GSM8K}} &  \textbf{Olympiad} &  \textbf{Minerva} & \multirow{2}{*}{\textbf{Avg.}} \\
 & & & &  &  \textbf{Bench} &  \textbf{Math} \\
\midrule
 1B & - & - & 31.4 & 54.0 & 7.7 &  6.2& 24.8\\
\midrule
\multicolumn{7}{c}{\textit{General Model, Base: Llama-3.2, RM: Llama3.1-8B-ORM-Deepseek-Data}} \\
\midrule
 1B & 7B & $N=4$ & 38.2 & 68.7	& 10.1 & 10.3 & 31.8 \\
 1B & 7B & $N=8$ & 43.2 & 74.7	& 12.4 & 10.3 & 35.2 \\
\cmidrule(lr){1-8}
\multicolumn{8}{c}{\textit{Question Truncation}} \\
 1B & 7B & $N=4$ & 37.0 & 62.8 & 9.8	& 9.6 & 29.8 \\
 1B & 7B & $N=8$ & 40.8 & 68.6	& 9.6 & 9.6	& 32.2 \\
\multicolumn{8}{c}{\textit{Initial Steps Truncation}} \\
 1B & 7B & $N=4$ & 35.2 & 62.0 & 9.2 & 11.4 & 29.5 \\
 1B & 7B & $N=8$ & 41.0 & 66.6	& 11.0 & 10.3 & 32.2 \\
\multicolumn{8}{c}{\textit{Last Steps Truncation}} \\
 1B & 7B & $N=4$ & 36.2 & 64.2	& 9.9 & 11.8 & 30.5 \\
 1B & 7B & $N=8$ & 37.6 & 68.8	& 10.4 & 13.6 & 32.6 \\
\multicolumn{8}{c}{\textit{Question Shuffling}} \\
 1B & 7B & $N=4$ & 29.0 & 54.7	& 8.6 & 9.2	& 25.4 \\
 1B & 7B & $N=8$ & 28.4 & 53.8	& 7.0 &	6.6	& 24.0 \\
\midrule
\multicolumn{8}{c}{\textit{General Model, Base: Llama-3.2, RM: Llama3.1-8B-PRM-Deepseek-Data}} \\
\midrule
 1B & 7B & $N=4$ & 35.4 & 64.1	& 9.8 & 8.5 & 29.5\\
 1B & 7B & $N=8$ & 41.2 & 69.1 & 10.4 & 11.4 & 33.0\\
\cmidrule(lr){1-8}
\multicolumn{8}{c}{\textit{Question Truncation}} \\
 1B & 7B & $N=4$ & 36.4 & 62.2	& 9.8 & 8.1 & 29.1\\
 1B & 7B & $N=8$ & 39.0 & 64.6 & 10.8 & 9.9 & 31.1\\
\multicolumn{8}{c}{\textit{Initial Steps Truncation}} \\
 1B & 7B & $N=4$ & 33.2 & 63.3	& 8.3 & 9.6 & 28.6\\
 1B & 7B & $N=8$ & 39.4 & 62.7	& 12.0 & 9.2 & 30.8\\
\multicolumn{8}{c}{\textit{Last Steps Truncation}} \\
 1B & 7B & $N=4$ & 33.6 & 62.5	& 8.1 & 10.3 & 28.6\\
 1B & 7B & $N=8$ & 36.4 & 64.9	& 7.7 & 9.9 & 29.7\\
\multicolumn{8}{c}{\textit{Question Shuffling}} \\
 1B & 7B & $N=4$ & 27.4 & 53.1 & 6.7 & 5.1 & 23.1\\
 1B & 7B & $N=8$ & 29.8 & 50.0 & 6.8 & 6.2 & 23.2\\
\bottomrule
\end{tabular}
\end{center}
\vskip -0.2in
\end{table*}

\subsection{Best-of-$N$}
Table~\ref{tab:qwen} and Table~\ref{tab:llama} present the Best-of-$N$ results under different input modifications for reward models. For the Qwen2.5-Math-Instruct-7B models, removing the question has a relatively minor impact on performance, with scores remaining stable across different values of $N$. For instance, when $N$=16, the model achieves an average score of 66.1, only 0.4 points lower than the vanilla Best-of-$N$ setting. This finding aligns with previous observations that reward models primarily rely on reasoning steps rather than the problem statement itself. In contrast, truncating initial or final steps leads to a more noticeable performance drop, highlighting the importance of complete reasoning trajectories. Additionally, question shuffling results in the largest performance degradation across both reward models, reinforcing the necessity of consistency between the question and solution for effective ranking.

For the Llama-3.2 base model, we evaluate both ORM and PRM reward models. Similar to the Qwen models, removing the question leads to only minor performance degradation. However, the most significant drop is observed in the question shuffling condition, particularly in Olympiad Bench and Minerva Math, where the disrupted semantic alignment prevents the reward model from identifying high-quality reasoning trajectories among multiple samples. With shuffled questions, the average performance drops to 24.0, even worse than the baseline of 24.8. These results further emphasize that reward models prioritize structural consistency and reasoning flow over explicit problem comprehension.

Best-of-$N$ results reflect the performance of the trajectory assigned the highest reward score, offering insights into how different input modifications impact the reward model’s ability to identify the best solution. The results in the table reinforce the key finding that reward models prioritize structural consistency over causal understanding. The relatively stable performance in question truncation suggests that the reward model does not rely on the problem statement itself to evaluate responses. Instead, it primarily assesses the internal coherence of the solution trajectory. However, the consistency between the question and the solution remains crucial, as disruptions can significantly degrade performance, highlighting the importance of semantic alignment in reward-based ranking.

\section{Discussion and Conclusion}
In this paper, we investigate the impact of input modifications on reward models and uncover key insights into their evaluation behavior. Our findings reveal that truncating the question has minimal impact on both absolute reward values and ranking consistency, suggesting that reward models primarily evaluate the solution trajectory rather than the problem statement itself. However, shuffling the question or modifying numerical values significantly alters the reward model’s output, indicating that semantic coherence and numerical consistency play a crucial role in assessment. Additionally, incomplete reasoning steps lead to substantial changes in rankings, highlighting the model’s strong reliance on a structured and complete reasoning trajectory rather than an explicit understanding of problem-solving steps.

\textbf{The Consistency Bias in Reward Models.} Our findings suggest that reward models are not truly evaluating the causal relationship between the question and its solution but rather the internal consistency of the reasoning process. Even when the question is removed, if the solution remains well-structured, the model continues to assign high scores. Conversely, when reasoning steps are truncated, the model struggles to maintain ranking consistency, indicating that it relies on pattern recognition rather than an actual causal understanding of the problem-solution relationship. This raises concerns about the model’s ability to generalize beyond familiar solution structures and adapt to novel problem distributions. Future research should explore techniques to mitigate this consistency bias and encourage a more causally grounded evaluation framework.

\textbf{Towards Causality-Aware Reward Models.}
To move beyond consistency-driven ranking, future reward models should incorporate causal reasoning techniques to better assess the logical validity of solutions. Potential directions include:
\begin{itemize}
    \item \textbf{Causality-Augmented Training}: Incorporating counterfactual reasoning tasks to train reward models to recognize causal dependencies rather than relying solely on surface-level patterns.
    \item \textbf{Chain-of-Thought Awareness}: Rewarding models not only for correct final answers but also for their adherence to logically structured reasoning chains, ensuring that each step contributes meaningfully to the solution.
    \item \textbf{Human-in-the-Loop Refinement}: Leveraging human preference data to penalize superficial pattern matching and encourage robust causal reasoning, improving the model’s ability to distinguish valid reasoning from plausible but incorrect trajectories.
\end{itemize}

\textbf{Reconsidering Reward Model Objectives.} Current reward models might be optimizing for ranking stability rather than true problem-solving ability. This raises the need to rethink selection strategies, such as:
\begin{itemize}
    \item Uncertainty-Aware Reward Models: Incorporating confidence-aware mechanisms to quantify the model’s uncertainty in evaluating complex reasoning tasks.
    \item Deeper Reasoning Signals: Designing reward functions that explicitly capture reasoning depth and logical validity, rather than solely relying on surface-level agreement with high-ranked answers.
\end{itemize}

\bibliography{myrefs}
\bibliographystyle{apalike}

\end{document}

%% file: myheaders.tex
\usepackage{amsmath,amsfonts,bm}

\def\eqref#1{Equation~(\ref{#1})}

\def\1{\bm{1}}

\RequirePackage{amsmath}
\RequirePackage{amssymb}
\RequirePackage{amsthm}
\RequirePackage{bm} 
\RequirePackage{url}
\usepackage{multirow}
\usepackage{natbib}
\usepackage{graphicx}
\usepackage{subfigure}
\usepackage{makecell}
\usepackage{booktabs}
\usepackage{array}
\usepackage{url}
\usepackage{algorithm}
\usepackage{algorithmic}
\usepackage{dsfont}

\usepackage{enumitem}

\usepackage{enumerate}
\usepackage[OT1]{fontenc}
\usepackage{natbib}

\usepackage{mathrsfs}

\usepackage{xcolor}
\usepackage{hyperref}

\def\##1\#{\begin{align}#1\end{align}}
\def\$#1\${\begin{align*}#1\end{align*}}

\usepackage{xcolor}

\definecolor{red1}{HTML}{f47983}
\definecolor{blue1}{HTML}{3eede7}
\definecolor{yellow1}{HTML}{f5dd6f}

